\documentclass[10pt,twocolumn,letterpaper]{article}

\usepackage[pagenumbers]{cvpr} % To force page numbers, e.g. for an arXiv version
\usepackage{tabularray}
\usepackage{multirow}
\usepackage{booktabs}
\usepackage{caption}
\usepackage{graphicx}
\usepackage{float}
\usepackage{subcaption}

\definecolor{cvprblue}{rgb}{0.21,0.49,0.74}
\usepackage[pagebackref,breaklinks,colorlinks,allcolors=cvprblue]{hyperref}

\def\paperID{30193} % *** Enter the Paper ID here
\def\confName{CVPR}
\def\confYear{2026}

\title{CoordSpeaker: Exploiting Gesture Captioning for Coordinated Caption-Empowered Co-Speech Gesture Generation}

\author{Fengyi Fang \qquad Sicheng Yang \qquad Wenming Yang \\
Tsinghua University\\
}

\begin{document}
\maketitle
\begin{abstract}
Co-speech gesture generation has significantly advanced human-computer interaction, yet speaker movements remain constrained due to the omission of text-driven non-spontaneous gestures (e.g., bowing while talking). 
Existing methods face two key challenges: 1) the semantic prior gap due to the lack of descriptive text annotations in gesture datasets, and 2) the difficulty in achieving coordinated multimodal control over gesture generation. 
To address these challenges, this paper introduces \textbf{CoordSpeaker}, a comprehensive framework that enables coordinated caption-empowered co-speech gesture synthesis. 
Our approach first bridges the semantic prior gap through a novel gesture captioning framework, leveraging a motion-language model to generate descriptive captions at multiple granularities. 
Building upon this, we propose a conditional latent diffusion model with unified cross-dataset motion representation and a hierarchically controlled denoiser to achieve highly controlled, coordinated gesture generation.
CoordSpeaker pioneers the first exploration of gesture understanding and captioning to tackle the semantic gap in gesture generation while offering a novel perspective of bidirectional gesture-text mapping.
Extensive experiments demonstrate that our method produces high-quality gestures that are both rhythmically synchronized with speeches and semantically coherent with arbitrary captions, achieving superior performance with higher efficiency compared to existing approaches. %Code and demo video are in the supplementary material and will be released upon paper acceptance. 
\end{abstract}    
\section{Introduction}
\label{sec:intro}

\begin{figure}[t] % [H]
\centering
\includegraphics[width=.96\linewidth]{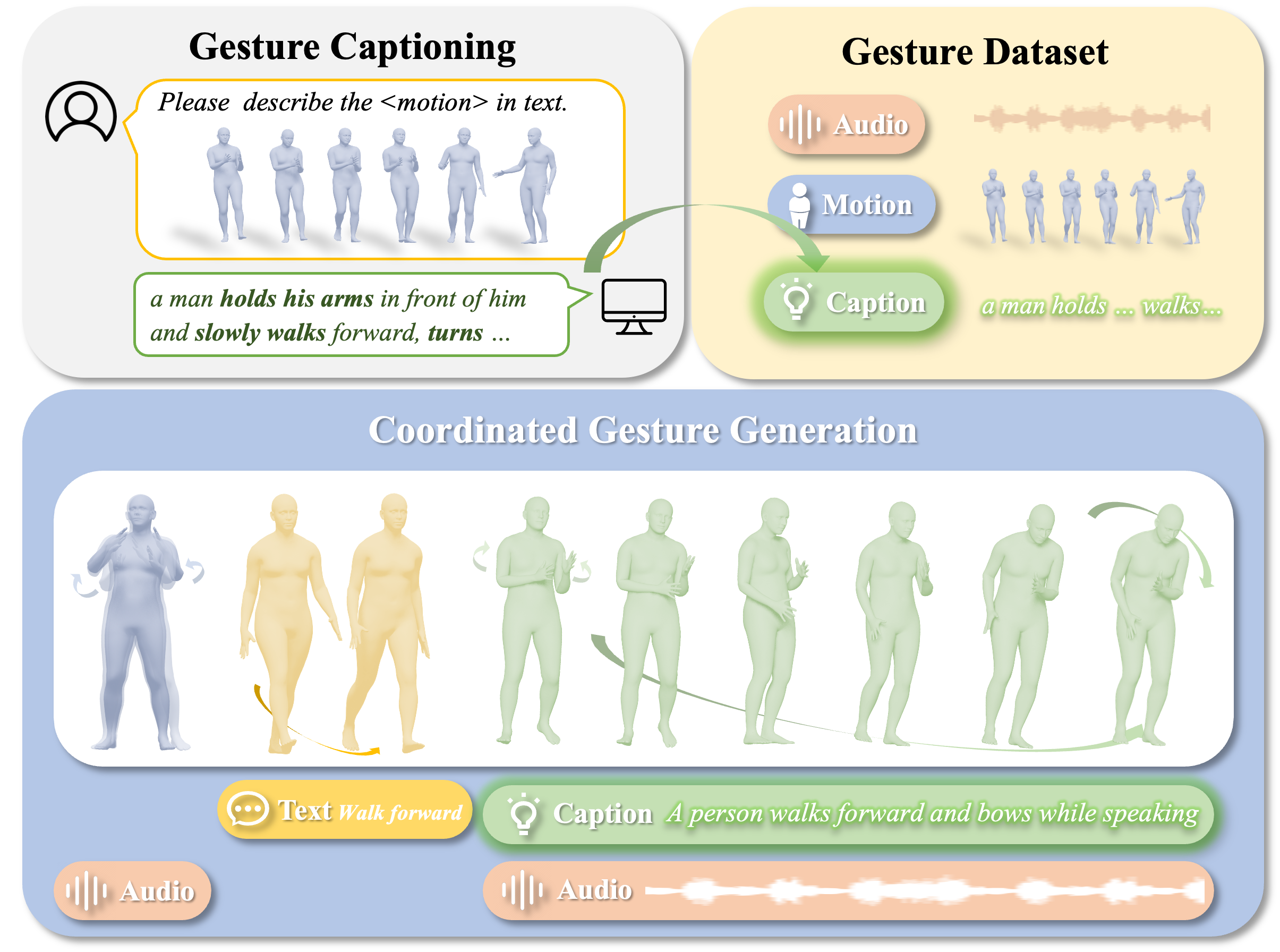} 
\caption{\textbf{CoordSpeaker} exploits \textit{gesture captioning} to enable customized \textit{coordinated speaker gesture generation}, producing both co-speech spontaneous gestures and caption-driven non-spontaneous motions.
In a speech scenario, our method allows the speaker to naturally walk forward and bow while speaking, seamlessly delivering a closing gesture.} 
\label{fig:teaser}
\end{figure}

Gesture synthesis has garnered significant interest due to its broad applications in human-computer interaction, such as virtual reality \cite{ahuja2019language2pose}, games \cite{zhu2023human}, and digital avatars \cite{liu2024emage,guo2022generating}.
To enhance the diversity and controllability of gesture generation, various modalities have been explored, including speech audio~\cite{yang2023diffusestylegesture, yang2023unifiedgesture,xu2025mambatalk}, text transcripts~\cite{zhi2023livelyspeaker, pang2023bodyformer,liu2024emage}, emotion~\cite{qi2024emotiongesture, qi2024weakly}, style~\cite{ao2023gesturediffuclip, yang2023diffusestylegesture, ghorbani2023zeroeggs}, and speaker identity~\cite{yang2023diffusestylegesture+}.
Among these, two critical aspects are highly emphasized: speech synchronization and semantic correlation.
Though advances in deep neural networks \cite{liu2024emage, xu2025mambatalk, cheng2024siggesture} have significantly enhanced generation quality and diversity, current methods primarily focus on co-speech spontaneous gestures, while neglecting text-driven non-spontaneous gestures \cite{yang2024freetalker}. 
This constrains the natural full-body movement of speakers, which is particularly critical in digital avatars applications like public speaking or gaming.
For instance, a game NPC may pace or lift an object while speaking, whereas a virtual teacher would naturally deliver lectures (spontaneous gesture) and point to key content on the screen as needed (non-spontaneous gesture).
Thus, achieving coordinated gesture generation that ensures rhythmic synchronization for spontaneous gestures and semantic coherence for non-spontaneous gestures is essential.

Coordinated gesture generation remains challenging due to two key factors: % applying these methods to co-speech gesture generation 
1) \textbf{Semantic Prior Gap}: Existing gesture datasets~\cite{liu2024emage,liu2022beat,ghorbani2023zeroeggs,talkingwithhands} lack direct descriptive text annotations. While speech audio is naturally available, semantic cues are typically inferred through speech transcriptions, which contain only limited and indirect guidance, resulting in weak semantic associations.
Moreover, manually annotating gesture semantics is prohibitively expensive, further exacerbating the challenge.
Gestures inherently possess richer semantic properties, which require a more direct and structured approach to extract and utilize.
Although some methods attempt to alleviate this issue by incorporating additional motion datasets for joint training \cite{yang2024freetalker} or motion-text alignment pretraining \cite{chen2024syntalker}, they still struggle with the inherent semantic gap.
The former fails to achieve joint control, while the latter introduces additional training and inference costs.
2) \textbf{Coordinated Multimodal Control Challenge}: %Coordinated, Synergistic, Joint
A fundamental challenge in generation domain lies in the effective coordination of heterogeneous multimodal conditions~\cite{zhang2024c3net,mughal2024convofusion,xu2024chain}, particularly when they impose distinct or conflicting control objectives. Unlike naturally paired modalities (e.g., speech-text transcription), combining indirectly related ones (e.g., audio and description) requires more carefully balancing conflicts and complementarities.
Previous methods simply concatenate multimodal conditions \cite{yang2024freetalker,chen2024syntalker,xu2025mambatalk}, treating them merely as signals of varying intensities rather than modeling their interactions, typically leading to inharmonious control.
Given these challenges, it is critical to bridge the semantic gap and develop a coordinated gesture generation method that enables harmonious multimodal control with minimal cost.

To tackle these challenges, we propose \textit{CoordSpeaker} (Fig.~\ref{fig:teaser}), a novel coordinated gesture generation approach that achieves both rhythmically synchronized and semantically consistent gesture synthesis while maintaining high generation quality and computational efficiency.
Firstly, to mitigate the semantic prior gap, we introduce a gesture captioning framework, utilizing a motion-language model to generate descriptive gesture captions and a multi-granular captioning mechanism to enable precise semantic guidance.
Secondly, to address the coordinated multimodal control challenge, we develop a coordinated gesture generation model learning a unified latent motion representation for cross-dataset modeling and leveraging a hierarchically controlled denoiser to ensure coordinated condition injection and efficient generation.
To the best of our knowledge, this work represents the first exploration of gesture captioning as a solution to semantic prior gap, facilitating coordination between speech and semantics in gesture generation, meanwhile offering a novel perspective for bidirectional gesture-text mapping.
Our contributions are as follows:
\begin{itemize}
    \item We propose CoordSpeaker, a coordinated gesture generation approach producing both co-speech spontaneous gestures and caption-driven non-spontaneous motions.
    \item We present the first gesture captioning framework to bridge the semantic prior gap of gesture data and ensure precise multi-granular caption alignment.
    \item We introduce a hierarchical conditional latent diffusion model enabling efficient and coordinated multimodal controlled gesture generation.
    \item Extensive experiments show our method effectively %generates descriptive gesture captions and 
    produces semantically coherent, rhythmically synchronized gestures while maintaining high fidelity and efficiency.
\end{itemize}

\section{Related Work}
\label{sec:relatedwork}

\subsection{Gesture Synthesis}
Gesture synthesis is a complex task that aims to generate gesture motion sequences from multimodal inputs~\cite{nyatsanga2023comprehensive}. 
Early approaches employed RNNs \cite{liu2022beat,yoon2020speech} and Transformers \cite{zhi2023livelyspeaker, pang2023bodyformer} for gesture sequence modeling, but suffered from limited temporal modeling capacity and computational challenges, respectively.
Diffusion models \cite{alexanderson2023listen,tevet2022mdm,zhang2024motiondiffuse} have gained prominence in motion generation due to powerful generative capabilities. However, operating directly in motion space leads to high computational costs and training instability. 
In comparison, latent diffusion models (LDMs) \cite{rombach2022high} enable powerful generative capabilities while maintaining computational efficiency through low-dimensional latent space operations.
Although current co-speech gesture generation methods have achieved significant progress in speech synchronization~\cite{chen2024diffsheg,zhu2023taming,zhang2023diffmotion}, semantic correlation is still limited by the lack of descriptive text annotations for gesture data~\cite{liu2024emage,liu2022beat}. 
Yang et al.~\cite{yang2024freetalker} propose a co-speech and text-driven approach, but it still suffers from semantic annotation gaps and does not support coordinated generation conditioned on joint signals.
Chen et al.~\cite{chen2024syntalker} leverage prompt-motion alignment pre-training to generate implicit text labels while introducing additional training and inference costs.
In contrast to existing methods, we propose a coordinated gesture generation approach, which enables both semantic correlation and rhythmic consistency while maintaining low annotation and computational costs.

\begin{figure*}[!t]
    \centering
    \includegraphics[width=.95\linewidth]{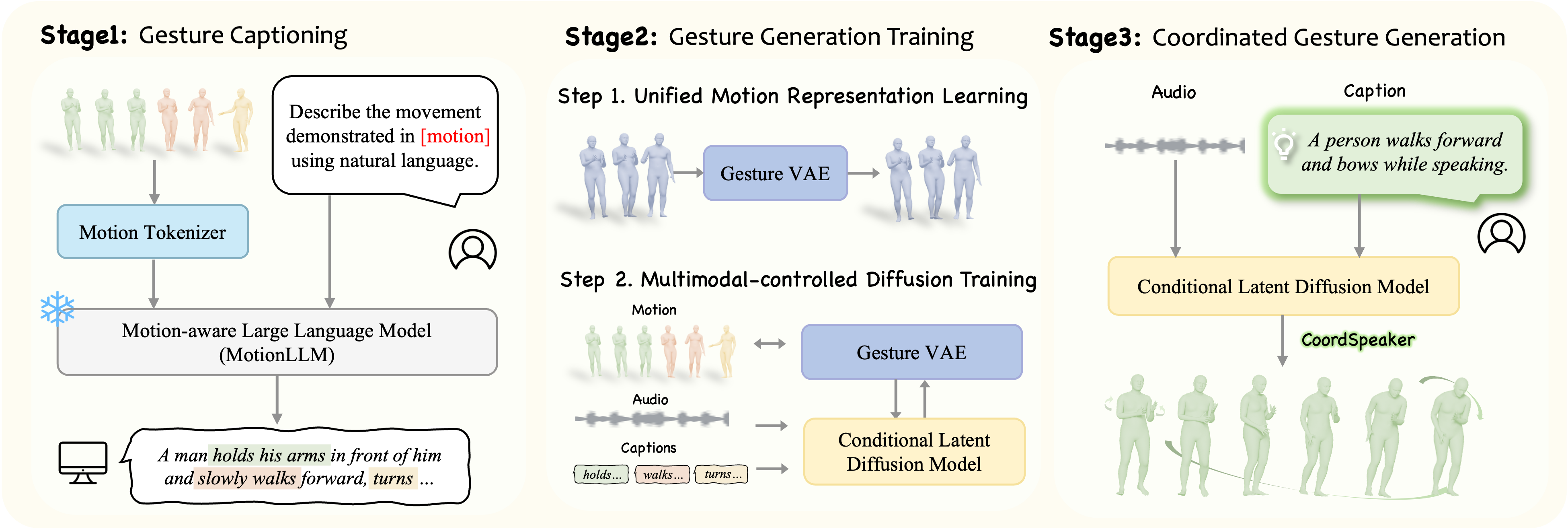}
   \caption{
    \textbf{Overview of CoordSpeaker.} 
    We first introduce Gesture Captioning (Sec.~\ref{sec:method:caption}) to bridge the semantic prior gap of gesture data, generating descriptive, multi-granular gesture captions at low cost.
    Subsequently, we propose a Coordinated Gesture Generation Model (Sec.~\ref{sec:method:generation}) enabling harmonious coordination over heterogeneous multi-modal and multi-scale conditions. % enable coordinated gesture generation through . 
    Our model can generate both rhythmically-synchronous and semantically-coherent gestures with high quality and superior efficiency.
    }
    \label{fig:method:pipeline}
\vspace{-6pt}
\end{figure*}

\subsection{Motion-Text Translation}
Human motion exhibits semantic coupling similar to natural language and is often viewed as a form of body language \cite{jiang2024motiongpt}. Previous research has explored various text-related motion tasks, including text-to-motion generation \cite{tevet2022mdm,guo2022humanml3d}, motion-to-text captioning \cite{jiang2024motiongpt,guo2022tm2t}, and unified motion-language modeling \cite{tevet2022motionclip,jiang2024motiongpt}.
Recent text-to-motion works \cite{chen2023executing,dai2024motionlcm} leverage pre-trained language models to extract semantics for motion control.
Motion captioning aims to describe human motion using natural language, with early approaches relying on statistical models and RNNs to learn motion-language mappings \cite{takano2015statistical}. 
More recently, bidirectional motion-text translation has gained attention.
TM2T \cite{guo2022tm2t} pioneered this direction through tokenization, %albeit within a single unified framework. 
unified motion-language models~\cite{jiang2024motiongpt,jiang2024motionchain,luo2024m} have emerged by fusing language data with large-scale motion models. 
Gestures inherently possess natural semantic properties, while no effort has been made to explore gesture understanding and captioning.
To address this gap, this paper first introduces a gesture captioning framework that bridges the lack of gesture captions and enables fine-grained semantic control over non-spontaneous gestures.

%%%%%%%%%%%%%%%%%%%%%%%%%%%%%%%%%%%%%%%%%%%%%%%%%%%%%%%%%%%%%%%%%%%%%%%%%%%%%%%%%%%%%%%%%%%%%%%%%%

\begin{figure*}[t]
    \centering
    \includegraphics[width=.9\linewidth]{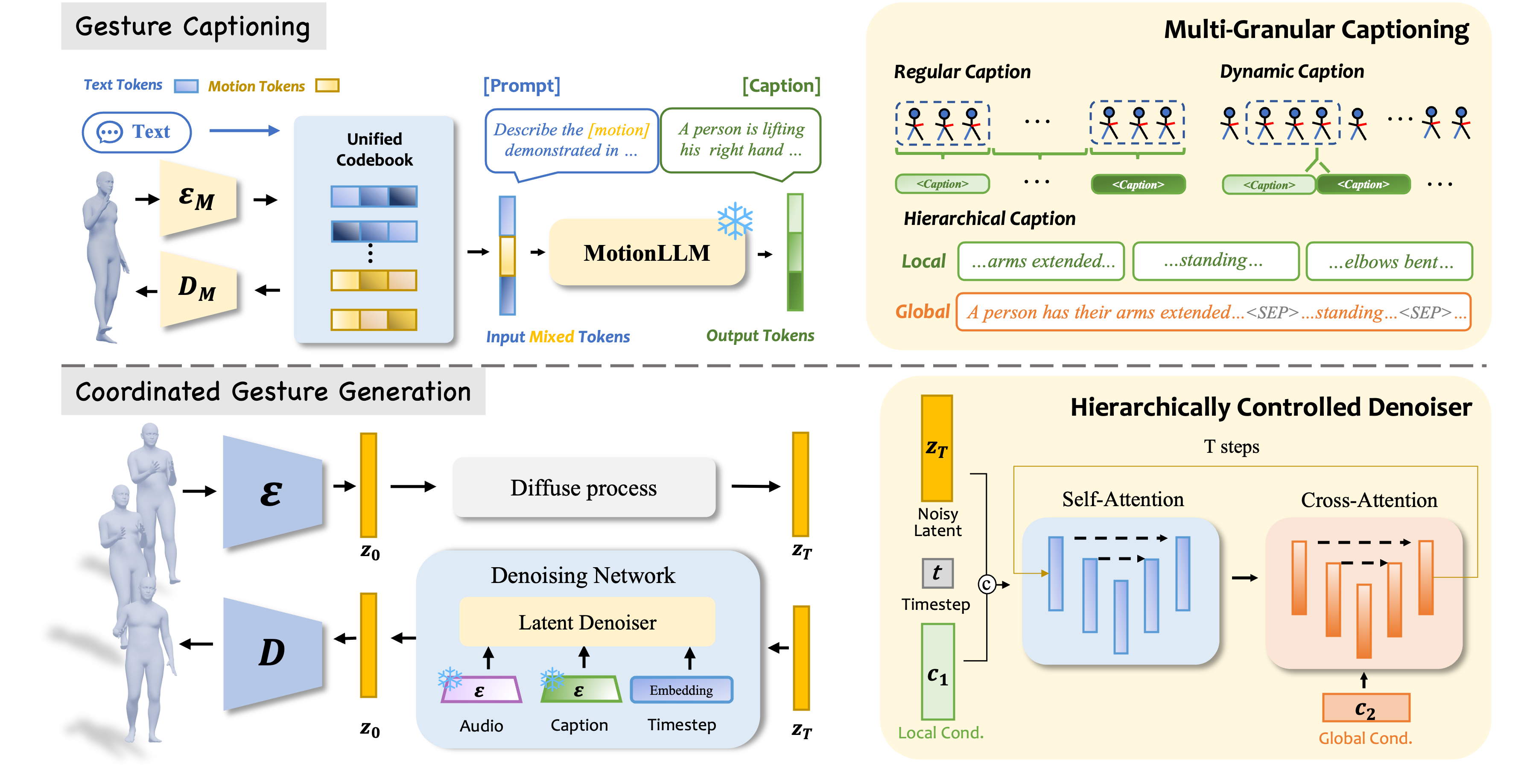}
\vspace{-4pt}
   \caption{
    Model overview. 
    \textbf{(Top) Gesture Captioning Framework:} A motion tokenizer and a motion-aware language model (MotionLLM) generate descriptive gesture \textit{[captions]} from predefined \textit{[prompt]} and \textit{[motion]} inputs, addressing the semantic prior gap efficiently. A multi-granular captioning mechanism further enhances multi-scale semantic alignment via three strategies: Regular, Dynamic, and Hierarchical.
    \textbf{(Bottom) Coordinated Gesture Generation Model:} A gesture VAE first learns a unified latent motion space for cross-dataset modeling. A conditional latent diffusion model with a hierarchically controlled denoiser enables efficient and coordinated gesture generation via hierarchical multimodal condition injection.}
    \label{fig:method:model}
\end{figure*}
\section{Method}
\label{sec:method}

In this section, we present a comprehensive framework for coordinated caption-empowered co-speech gesture generation (Fig.~\ref{fig:method:pipeline},~\ref{fig:method:model}). 
We first introduce a Gesture Captioning Framework (Sec.~\ref{sec:method:caption}) to bridge the semantic prior gap, leveraging a motion-language model combined with a multi-granular captioning mechanism to generate descriptive, precisely aligned gesture captions, enabling fine-grained semantic injection over gesture generation.
Subsequently, we propose a Coordinated Gesture Generation Model (Sec.~\ref{sec:method:generation}) comprising two key components: 
(1) a gesture VAE that learns a unified low-dimensional latent representation, enabling compact cross-dataset motion modeling, and (2) a conditional latent diffusion model with a hierarchically controlled denoiser ensuring coordinated multimodal-controlled gesture generation. 
Our model can generate both rhythmically-synchronous and semantically-coherent speaker gestures with high quality and efficiency.

\subsection{Gesture Captioning}
\label{sec:method:caption}

\subsubsection{Motion-Language Modeling}
\label{sec:method:mllm}
Our gesture captioning framework (Fig.~\ref{fig:method:model}, Top) comprises two main components: a \textit{motion tokenizer} and a \textit{motion-aware large language model (MotionLLM)}.
The motion tokenizer is built upon the VQ-VAE architecture used in~\cite{guo2022tm2t,zhang2023t2mgpt}, consisting of an encoder $\mathcal{E_M}$ and a decoder $\mathcal{D_M}$ that generates discrete motion tokens.
The motion-aware language model adopts a transformer-based architecture with a unified text-motion vocabulary $\mathbf{V} = \{\mathbf{V_t}, \mathbf{V_m}\}$, enabling flexible joint modeling of text and motion within a single model.
To better match the general motion-language space, the gesture sequence is first converted into a unified motion representation space (detailed in Sec.~\ref{sec:method:m_rep}), then projected to a 22-joint feature subset. %fed into our gesture captioning framework.
Specifically, given an M-frame gesture motion sequence $m^{1:M}=\{x^i\}^M_{i=1}$, the motion tokenizer encodes and quantizes it into a discrete motion token sequence $s^{1:L}=\{s^i\}^L_{i=1}$. A carefully designed \textit{prompt template} is then tokenized into text tokens $w^{1:N}=\{w^i\}^N_{i=1}$. Subsequently, the discrete motion and text tokens are mixed and jointly fed into the motion-aware language model to generate the corresponding gesture caption $\hat{w}^{1:L}=\{w^i\}^L_{i=1}$.
In practice, we leverage MotionGPT~\cite{jiang2024motiongpt} for motion-language modeling and freeze it during inference.
Notably, the captions are generated and cached offline, incurring no additional overhead. This framework produces direct and descriptive captions for gesture data, efficiently addressing the semantic prior gap, enabling precise semantic control over gesture generation.
See supplementary material (Sec.~\ref{sec:appendix_cap}) for more details.

\subsubsection{Multi-Granular Captioning}
\label{sec:method:control}

Precise semantic control via gesture captions is challenging due to temporal dynamics and varying semantic granularity. To address this, we propose a \textit{multi-granular captioning} mechanism that enables fine-grained caption alignment across temporal and semantic scales.
We first introduce a \textit{hierarchical} caption manner integrating both local and global gesture captions. 
Given a gesture sequence, we segment it as $m^{1:M}=\{m^{i:i+K-1}\}_{i=1}^{M-K+1}$, where $K$ denotes the segment length. Each segment is processed by our gesture captioning framework independently to generate \textit{local} captions $\mathbf{C}_\text{local}=\{w^i\}^L_{i=1}$. 
In contrast, \textit{global} caption is formed by concatenating local captions with separator tokens $\mathbf{C}_\text{global}=\{\mathbf{C}^1_\text{local} \textless\text{SEP}\textgreater \mathbf{C}^2_\text{local} \textless\text{SEP}\textgreater \dots\mathbf{C}^{M/K}_\text{local}\}$. %\textit{\textless SEP\textgreater}. 
While local captions provide fine-grained semantic supervision at the segment level, a global caption summarizes the overall gesture semantics across the entire sequence.
To instantiate this mechanism, we explore three captioning strategies (Fig.\ref{fig:method:model}, Top):
(1) \textit{Regular Caption} applies uniform temporal segmentation to generate local captions for each fixed-length gesture segment, offering precise temporal alignment. 
(2) \textit{Dynamic Caption} randomly samples segments during training, and introduces stochastic sampling and caption mixing for local captions, enhancing the robustness to varying temporal patterns and flexible caption combinations.
(3) \textit{Hierarchical Caption} combines both local and global captions to incorporate complementary fine-to-coarse semantic cues. %to enrich captions with multi-scale descriptive information.
This multi-granular captioning mechanism ensures precise caption alignment and enables flexible multi-scale semantic control over gesture generation.

\subsection{Coordinated Gesture Generation}
\label{sec:method:generation}
\subsubsection{Unified Motion Representation}
\label{sec:method:m_rep}

To leverage additional human motion semantic priors, we incorporate the human motion dataset HumanML3D~\cite{guo2022humanml3d} and introduce a \textit{unified motion representation}. %to encode diverse motion data into a consistent feature space. 
The various data formats are first converted into a unified feature format and then fit into a representative latent space via our motion encoder $\mathcal{E}$ (Fig.~\ref{fig:method:model}).
Specifically, data in different formats is first converted into SMPL-X~\cite{pavlakos2019expressive} axis-angle representation. After scaling and initial orientation adjustment, 3D joint positions are obtained through SMPL-X forward computation.
Subsequently, following~\cite{yang2024freetalker,guo2022humanml3d}, the $i$-th motion frame is represented as $x^i=\left\{\dot{r}^a, \dot{r}^x, \dot{r}^z, r^y, \mathbf{j}^p, \mathbf{j}^v, \mathbf{j}^r, \mathbf{c}^f\right\}$, where $\dot{r}^a, \dot{r}^x, \dot{r}^z, r^y$ denote the root joint's angular velocity, linear velocities and height, $\mathbf{j}^p, \mathbf{j}^v, \mathbf{j}^r$ represent joint positions, velocities and rotations, and $\mathbf{c}^f$ indicates foot contact.
We employ 55 joints to accommodate gesture data better. Consequently, each motion frame is represented by a 659-dimensional feature vector, denoted as $x \in \mathbb{R}^{T\times 659}$, where $T$ is the sequence length.

\subsubsection{Gesture VAE} 
A transformer-based~\cite{petrovich2021action} VAE model (Fig.~\ref{fig:method:model}) is adopted to encode motion representation into a compact and informative latent space, which consists of an encoder $\mathcal{E}$ and a decoder $\mathcal{D}$, both enhanced with long skip connections to preserve motion details. %~\cite{ronneberger2015u} 
Specifically, the motion sequence $x^{1:L}$ is first encoded into a latent vector $z \in \mathbb{R}^{n\times d}$ through the encoder $\mathcal{E}$, where $d$ denotes the latent dimension. The encoder takes frame-wise motion features and learnable distribution tokens as input, producing Gaussian distribution parameters $\mu$ and $\sigma$ for the motion latent space. These parameters are used to reparameterize~\cite{kingma2013auto} the latent vector through the standard VAE sampling process.
For motion decoding, $\mathcal{D}$ employs a cross-attention~\cite{vaswani2017attention} mechanism. It takes zero motion tokens as queries and the latent vector $z$ as key and value, generating the reconstructed motion sequence $\hat{x}^{1:L}$. The VAE is trained by minimizing a combination of Mean Squared Error (MSE) for reconstruction accuracy and Kullback-Leibler (KL) divergence to regularize the encoded latent distribution $q(z|x^{1:L}) = \mathcal{N}(z; \mu_{\mathcal{E}}, \sigma^2_{\mathcal{E}})$ toward a standard gaussian distribution:
\begin{equation}
\mathcal{L}_{\text{VAE}} = \|x^{1:L} - \hat{x}^{1:L}\|_2^2 + \beta \text{KL}(q(z|x^{1:L}) \| p(z))
\label{eq:loss_vae}
\end{equation}
where $\beta$ balances the reconstruction and regularization terms. This latent representation enables efficient gesture synthesis while maintaining motion fidelity and diversity.

\subsubsection{Conditional Latent Diffusion Model}
\label{sec:method:gld}
We introduce a \textit{conditional latent diffusion model} with a hierarchically controlled denoiser (Fig.~\ref{fig:method:model}, Bottom) to generate high-quality, coordinated gestures in learned latent space.

\paragraph{Diffuse Process. }
The diffusion process in latent space follows a Markov chain that gradually adds Gaussian noise to the latent vector $z \in \mathbb{R}^{n\times d}$. For a noising step $t \in [1,T]$, the forward process is defined as:
\begin{equation}
    q(z_t|z_{t-1}) = \mathcal{N}(\sqrt{\alpha_t}z_{t-1}, (1-\alpha_t)\mathbf{I}),
\end{equation}
$\alpha_t \in (0,1)$ is constant variance schedule, $z_T \sim \mathcal{N}(0, \mathbf{I})$.

\paragraph{Hierarchically Controlled Denoiser. } 
We propose a hierarchically controlled denoiser $\epsilon_\theta$ to facilitate coordinated multimodal control over gesture generation.
The denoiser employs a transformer-based endoder-decoder architecture to predict and remove noise iteratively. Starting from random noise $z_T$, the denoiser gradually recovers the latent vector $\hat{z}_0$, which is then decoded into gesture motion through $\mathcal{D}$. 
Specifically, given the caption embedding $\mathbf{C}$ and audio embedding $\mathbf{A}$,
the denoising process is controlled by \textit{hierarchical two-stage condition injection} (Fig.~\ref{fig:method:model}):
First, the \textit{local} condition $\mathbf{c_1}=\{\mathbf{C}_\text{local}, \mathbf{A}\}$ is concatenated with the noised latent and timestep embeddings before being fed into the denoiser encoder $\mathcal{E}_d$ performing self-attention, ensuring precise semantic and rhythmic synchronization with the local gesture segments. %providing basic guidance.
Second, the \textit{global} condition $\mathbf{c_2}=\{\mathbf{C}_\text{global}\}$ is incorporated into the denoiser decoder $\mathcal{D}_d$ through cross-attention, enhancing both the overall gesture coherence and semantic relevance with high-level context. %enabling fine-grained control tuning with high-level context.
Overall, the reverse process is defined as follows:
\begin{equation}
\label{eq:hie_cond}
    h = \mathcal{E}_d(\text{concat}(z_t, t, \mathbf{c_1})), \quad
    \epsilon_\theta(z_t, t, \mathbf{c}) = \mathcal{D}_d(h, \mathbf{c_2}),
\end{equation}
where $t$ is the timestep embedding, %$\mathbf{C}$ and $\mathbf{A}$ denote caption and audio embeddings, 
and the predicted noise $\epsilon_\theta$ is used to recover $\hat{z}_{t-1}$.
This hierarchical architecture aligns with our multi-granular captions effectively coordinates different modalities and scales, ensuring balanced contributions for heterogeneous conditions.

\paragraph{Classifier-free Guidance. }
To enhance the quality and controllability of generated gestures, we adopt classifier-free guidance \cite{ho2022classifier}. During training, 10\% of the condition inputs are randomly masked~\cite{saharia2022photorealistic} to learn both conditional and unconditional distributions. During inference, noise prediction is computed as a weighted combination of outputs with different guidance:
\begin{equation}
    \begin{split}
        \epsilon_\theta^s(z_t, t, \mathbf{c}) = & s_1 \epsilon_\theta(z_t, t, \mathbf{c}=\{\varnothing, \mathbf{A}\}) \\
        & + s_2 \epsilon_\theta(z_t, t, \mathbf{c}=\{\mathbf{C}, \varnothing\}) \\
        & + (1 - s_1 - s_2) \epsilon_\theta(z_t, t, \mathbf{c}=\{\varnothing, \varnothing\}),
    \end{split}
\end{equation}
where %$\mathbf{C}$ and $\mathbf{A}$ denote caption and audio embeddings, and 
$s_1$, $s_2$ are guidance scales for audio $\mathbf{A}$ and caption $\mathbf{C}$, respectively.

\paragraph{Training and Inference. }
The latent diffusion model is trained with a $\ell_2$ objective \cite{ho2020ddpm}: 
\begin{equation} 
    \mathcal{L}_{\text{Diff}} = \mathbb{E}_{\epsilon,t}[\|\epsilon - \epsilon_\theta(z_t, t, \mathbf{c})\|_2^2],
\end{equation}
where $\epsilon \sim \mathcal{N}(0,\mathbf{I})$ and $z_0 = \mathcal{E}(x^{1:L})$. During inference, our model employs DDIM sampler~\cite{song2020ddim} with 50 denoising steps to predict the latent $\hat{z}_0$, then decodes it to motion sequence $\hat{x}^{1:L}$ using a forward pass through decoder $\mathcal{D}$.

\begin{figure*}[t]
    \centering
    \includegraphics[width=.9\linewidth]{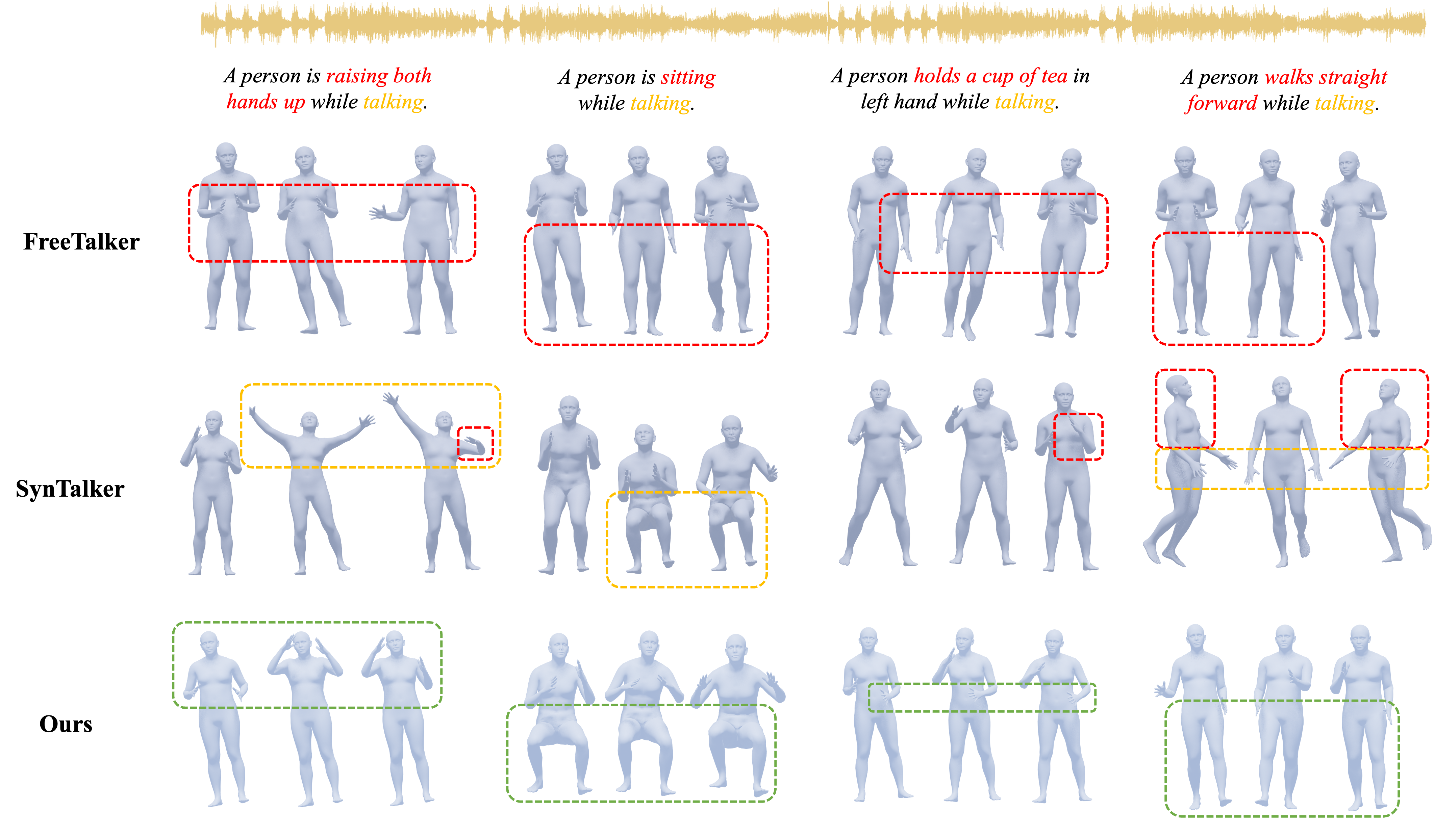}
\vspace{-5pt}
    \caption{Qualitative comparison of coordinated gesture generation. \textcolor{red}{Red} boxes highlight semantic inconsistencies, \textcolor[RGB]{204,153,0}{yellow} boxes indicate unnatural motions, and \textcolor[RGB]{34,139,34}{green} boxes denote well-coordinated natural gestures. More results are in supplementary material (Sec.~\ref{sec:appendix_visual}).} 
    \label{fig:method:visualization}
\end{figure*}

\begin{table*}[t]
    \centering
    \caption{Quantitative results of baseline comparisons and ablation studies. Metrics are reported with 95\% confidence interval over 20 runs. `$\rightarrow$' denotes the closer to the real motion the better. We report BC $\times 10^{-1}$ and Top-1 R-Precision.}
    \small
    \label{tab:main_results}
    \begin{tabular}{l c@{\hspace{3pt}} c@{\hspace{3pt}} c@{\hspace{3pt}} c@{\hspace{3pt}} c@{\hspace{3pt}} c@{\hspace{3pt}} c@{\hspace{3pt}} c@{\hspace{3pt}} c}
    \toprule
    \multirow{3}{*}{Methods} & \multicolumn{2}{c}{Reconstruction} & \multicolumn{3}{c}{Audio-to-Gesture} & \multicolumn{4}{c}{Text-to-Motion} \\
    \cmidrule(lr){2-3} \cmidrule(lr){4-6} \cmidrule(lr){7-10}
    & Jerk$\rightarrow$ & Accel.$\rightarrow$ & FGD$\downarrow$ & BC$\uparrow$ & L1Div$\uparrow$ & FID$\downarrow$ & MM-Dist$\downarrow$ & Div$\rightarrow$ & R-Precision$\uparrow$ \\
    \midrule
    GT & $1.165^{\pm .000}$ & $0.043^{\pm .000}$ & - & - & - & - & $6.205^{\pm .043}$ & $5.512^{\pm .114}$ & $0.140^{\pm .008}$ \\
    Freetalker & $0.611^{\pm .013}$ & $0.030^{\pm .000}$ & $2.101^{\pm .026}$ & $1.147^{\pm .028}$ & $11.332^{\pm .025}$ & $0.761^{\pm .048}$ & $6.737^{\pm .051}$ & $5.396^{\pm .127}$ & $0.102^{\pm .008}$ \\
    % 【Hie_cap + global_cross】
    Ours  & $1.190^{\pm .015}$ & $0.039^{\pm .001}$ & $3.173^{\pm .123}$ & $1.327^{\pm .049}$ & $10.861^{\pm .066}$ & $1.118^{\pm .061}$ & $6.814^{\pm .056}$ & $5.558^{\pm .126}$ & $0.100^{\pm .008}$ \\
    \midrule
    Ours-w/o hcd. & $1.201^{\pm .017}$ & $0.038^{\pm .001}$ & $2.302^{\pm .061}$ & $1.910^{\pm .004}$ & $12.781^{\pm .044}$ & $1.260^{\pm .063}$ & $6.872^{\pm .058}$ & $5.303^{\pm .107}$ & $0.102^{\pm .010}$ \\
    Ours-w/o mgc. & $1.239^{\pm .013}$ & $0.039^{\pm .000}$ & $3.123^{\pm .139}$ & $2.256^{\pm .045}$ & $15.363^{\pm .103}$ & $2.568^{\pm .099}$ & $7.031^{\pm .044}$ & $5.447^{\pm .150}$ & $0.082^{\pm .006}$ \\
    Ours-w/o mo. & $1.005^{\pm .014}$ & $0.032^{\pm .000}$ & $2.654^{\pm .041}$ & $2.627^{\pm .043}$ & $19.600^{\pm .089}$ & $3.911^{\pm .163}$ & $7.664^{\pm .050}$ & $4.070^{\pm .117}$ & $0.043^{\pm .004}$ \\
    \bottomrule
    \end{tabular}
\end{table*}

\section{Experiments}
\label{sec:experiments}
In this section, we evaluate our method through both quantitative and qualitative analyses in four aspects:  
(1) \textbf{Coordinated Gesture Generation}. Assessing the model’s ability to generate speech-synchronized, semantically-relevant gestures under joint speech and caption control, compared to state-of-the-art baselines.
(2) \textbf{Text-Driven Motion Generation}.
Comparing with state-of-the-art text-to-motion methods to evaluate semantic understanding and non-spontaneous gesture generation. 
(3) \textbf{Gesture Captioning}.  Evaluating the quality, human-alignment, and diversity of generated captions marks the first approach to bridging the semantic gap in gesture datasets.
(4) \textbf{Ablation Studies}. Analyzing the impact of key components in our method.

\subsection{Experimental Setup}
\paragraph{Datasets}
We jointly train on the audio-to-gesture dataset BEAT~\cite{liu2022beat} and the text-to-motion dataset HumanML3D~\cite{guo2022humanml3d}.  %This cross-dataset learning strategy allows us to incorporate additional semantic motion priors.
BEAT offers 76 hours of speech-gesture data. We utilize four English speakers' gestures following~\cite{liu2022beat}. %covering diverse gesture types and speaking scenarios, 
% offering crucial support for speech-driven gesture generation. 
HumanML3D contains 14,616 motions with 44,970 text descriptions, providing rich semantic priors. 
For coordinated generation, missing text in BEAT is generated via our captioning framework, while missing audio in HumanML3D is set to zero following~\cite{yang2024freetalker}. 

% \vspace{-10pt}
\paragraph{Metrics}
The evaluation is conducted from three perspectives:
(1) Reconstruction Quality: Motion smoothness and naturalness are measured using jerk and acceleration~\cite{kucherenko2019analyzing}.
(2) Audio-to-Gesture: Gesture realism is assessed via FGD~\cite{yoon2020speech}, diversity by the mean L1 distance between gestures~\cite{liu2024emage}, and speech-motion synchronization by Beat Consistency (BC)~\cite{li2021ai}.
(3) Text-to-Motion: Frechet Inception Distance (FID)~\cite{chen2023executing} and Diversity (DIV)~\cite{guo2022generating} evaluate realism and diversity, while motion-retrieval precision (R Precision) and Multimodal Distance (MM-Dist) assess motion-text alignment~\cite{chen2023executing}. 
Implementation details are in supplementary material (Sec.~\ref{sec:appendix_imple}).

\subsection{Coordinated Gesture Generation}
\label{sec:exp_main}

% \vspace{-2pt}
\paragraph{Qualitative Comparison}
We first evaluate the performance of our method in generating coordinated, speech-synchronized, semantically-relevant gestures.
As depicted in Fig.~\ref{fig:method:visualization}, we compare with the only two works~\cite{yang2024freetalker,chen2024syntalker} that addressed related coordinated generation tasks. %visual analysis was performed to evaluate the coordinated generation performance of the proposed method.
Following~\cite{chen2024syntalker}, results are presented with a \textit{calm} audio input and four distinct text captions. %: \textit{``raising both hands up''}, \textit{``sitting''}, \textit{``holds a cup of tea''}, and \textit{``walks straight forward''}.
Results show that our method generates well-coordinated gestures that are both rhythmically aligned and semantically consistent, while achieving more natural appearances.
In contrast, FreeTalker~\cite{yang2024freetalker} fails to produce semantic motions under all settings, focusing solely on speech-driven gestures. 
While SynTalker~\cite{chen2024syntalker} captures partial semantic relevance, it struggles to jointly coordinate multimodal conditions, leading to control conflicts (e.g., losing co-speech gestures in \textit{``walks while talking''}), inconsistent details (e.g., incorrect hand poses for \textit{``raising''} and \textit{``holding''}), and unnatural stiffness (e.g., incorrect turning in \textit{``walks straight forward''}).
Ablations (Sec.~\ref {sec:exp_ablation}) further demonstrate that our superior semantic understanding stems from gesture captioning and enhanced multimodal coordination from proposed hierarchically controlled denoiser. More results are in supplement (Sec.~\ref{sec:appendix_visual} and video).

% \vspace{-10pt}
\paragraph{Quantitative Results}
Since only FreeTalker~\cite{yang2024freetalker} and SynTalker~\cite{chen2024syntalker} share a similar research scope with our work, and SynTalker does not provide a quantitative evaluation method under multimodal conditioning, we reproduce FreeTalker as primary baseline in this comparison, and present additional single-condition comparison with SynTalker in Sec.~\ref{sec:exp_t2m}.
As shown in Table~\ref{tab:main_results}, our method achieves comparable or superior results across all evaluation metrics, surpassing FreeTalker in reconstruction quality, beat consistency (BC $0.180\uparrow$), and diversity (Div $0.162\uparrow$), while achieving better coordination and substantially faster inference (Sec.~\ref {sec:exp_ablation}).
Note that due to existing quantitative metrics solely focus on single-modality factors, our aim is to achieve balance performance across multimodal metrics. More details are in supplement (Sec.~\ref{sec:appendix_metric}).

% \vspace{-10pt}
\paragraph{Perceptual Study}
A user study was conducted to assess gesture quality under joint speech and caption control. 20 participants were recruited to evaluate 10 pairs of 9-second results based on: (i) \textit{Naturalness}: realism and naturalness of generated gestures; (ii) \textit{Synchrony}: synchronization with speech; (iii) \textit{Matching}: alignment with text captions. Participants viewed video clips from different models and selected the best one for each aspect.
Table~\ref{tab:perceptual_study} shows our proposed model outperforms others. A chi-square test further confirms the significant differences across methods in all aspects (Naturalness: $\chi^{2}$=27.20, $p$=0.000005; Synchrony: $\chi^{2}$=8.68, $p$=0.034; Matching: $\chi^{2}$=17.72, $p$=0.0005).
More details are in supplementary material (Sec.~\ref{sec:appendix_userstudy}).

\begin{table}[t]
    \centering
    \caption{User preference win rates (\%) show that our results are perceived as more realistic and controllable, outperforming previous work~\cite{chen2024syntalker} by 4.0\%, 5.5\%, and 1.5\% in naturalness ($p{<}0.001$), synchrony ($p{<}0.05$), and matching ($p{<}0.001$), respectively. All results are reported with 95\% confidence intervals.}
    \small
    \label{tab:perceptual_study}
    \begin{tabular}{l ccc}
    \toprule
    Methods & Naturalness & Synchrony & Matching \\
    \midrule
    FreeTalker & $18.0^{\pm 5.32}$ & $24.0^{\pm 5.92}$ & $15.5^{\pm 5.02}$ \\ 
    SynTalker & $32.0^{\pm 6.46}$ & $26.5^{\pm 6.12}$ & $31.5^{\pm 6.44}$ \\ 
    Ours-w/o cap. & $14.0^{\pm 4.81}$ & $17.5^{\pm 5.27}$ & $20.0^{\pm 5.54}$ \\
    Ours & $36.0^{\pm 6.65}$ & $32.0^{\pm 6.46}$ & $33.0^{\pm 6.52}$ \\
    \bottomrule
    \end{tabular}
% \vspace{-8pt}
\end{table}

\begin{table*}[t]
    \centering
    \caption{Comparison with state-of-the-art methods on HumanML3D test set. Metrics follow the standard protocol in \cite{chen2023executing} and are reported with 95\% confidence interval over 20 runs. `$\rightarrow$' denotes the closer to the real motion the better.}
    \label{tab:h3d_results}
    \small
    \begin{tabular}{l cccccc}
    \toprule
    \multirow{2}{*}{Methods} & \multirow{2}{*}{FID$\downarrow$} & \multirow{2}{*}{MM-Dist$\downarrow$} & \multirow{2}{*}{Div$\rightarrow$} & \multicolumn{3}{c}{R-Precision$\uparrow$} \\
    \cmidrule(lr){5-7}
    & & & & Top-1 & Top-2 & Top-3 \\
    \midrule
    GT  &  $0.001^{\pm .001}$ & $3.378^{\pm .007}$ & $10.471^{\pm .083}$ &  $0.490^{\pm .003}$ & $0.682^{\pm .003}$ & $0.783^{\pm .003}$ \\
    MDM~\cite{tevet2022mdm} & $1.390^{\pm .088}$ & $4.599^{\pm .037}$ & $10.704^{\pm .066}$ & $0.363^{\pm .007}$ & $0.553^{\pm .008}$ & $0.662^{\pm .007}$ \\
    T2M-GPT~\cite{zhang2023t2mgpt} &  $0.564^{\pm .012}$ & $3.867^{\pm .008}$ & $10.558^{\pm .083}$  & $0.433^{\pm .003}$ & $0.615^{\pm .002}$ & $0.716^{\pm .003}$ \\
    MLD~\cite{chen2023executing} & $0.963^{\pm .029}$ & $3.898^{\pm .012}$ & $10.401^{\pm .096}$ & $0.429^{\pm .003}$ & $0.613^{\pm .003}$ & $0.717^{\pm .002}$ \\
    MoMask~\cite{guo2024momask} & $0.222^{\pm .007}$ & $3.620^{\pm .011}$ & $10.621^{\pm .096}$ & $0.461^{\pm .002}$ & $0.657^{\pm .003}$ & $0.760^{\pm .002}$ \\
    SynTalker~\cite{chen2024syntalker} & $4.385^{\pm .034}$ & $4.499^{\pm .012}$ & $9.374^{\pm .073}$ & $0.375^{\pm .003}$ & $0.564^{\pm .003}$ & $0.681^{\pm .002}$ \\
    \midrule
    Ours  & $0.405^{\pm .012}$ & $3.584^{\pm .012}$ & $9.109^{\pm .235}$  & $0.424^{\pm .003}$ &$0.601^{\pm .003}$ & $0.702^{\pm .003}$ \\
    \bottomrule
    \end{tabular}
\end{table*}

\subsection{Text-Driven Motion Generation}
\label{sec:exp_t2m}
We further compare our method with SOTA text-to-motion approaches to validate its capacity in capturing semantics and generating non-spontaneous motions. 
For fair comparison, we report results on the HumanML3D test set using only text condition.
Table~\ref{tab:h3d_results} shows our method achieves advanced performance compared to SOTA models, attaining the best text alignment (MM-Dist 3.584) and the second-best generation fidelity (FID 0.405). This underscores the effectiveness of our hierarchically controlled denoiser in integrating fine-grained semantics.
Notably, our approach significantly outperforms the similar synergistic method SynTalker, highlighting its superiority in both coordinated generation and enhanced semantic control.

\begin{figure*}[t]
    \centering
  \subcaptionbox{\label{fig:visualization_caption}}
    {\includegraphics[width=0.5\linewidth]{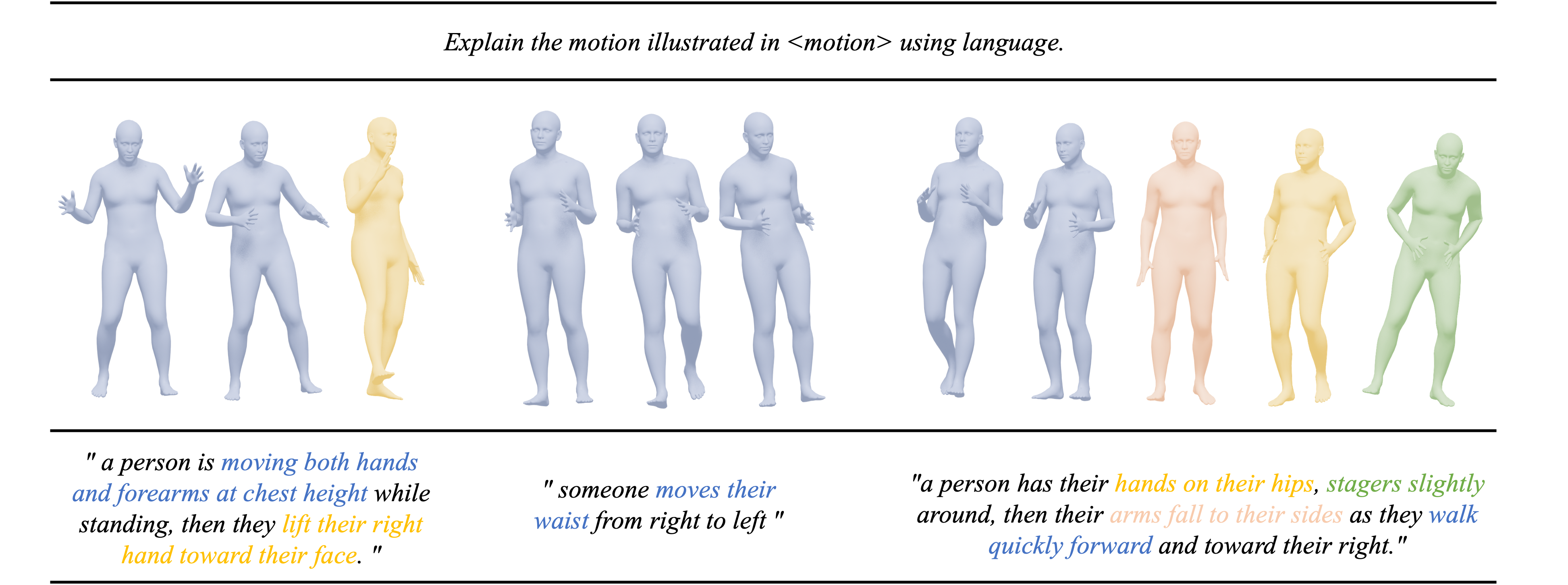}}
  \subcaptionbox{\label{fig:radar}}
    {\includegraphics[width=0.2\linewidth]{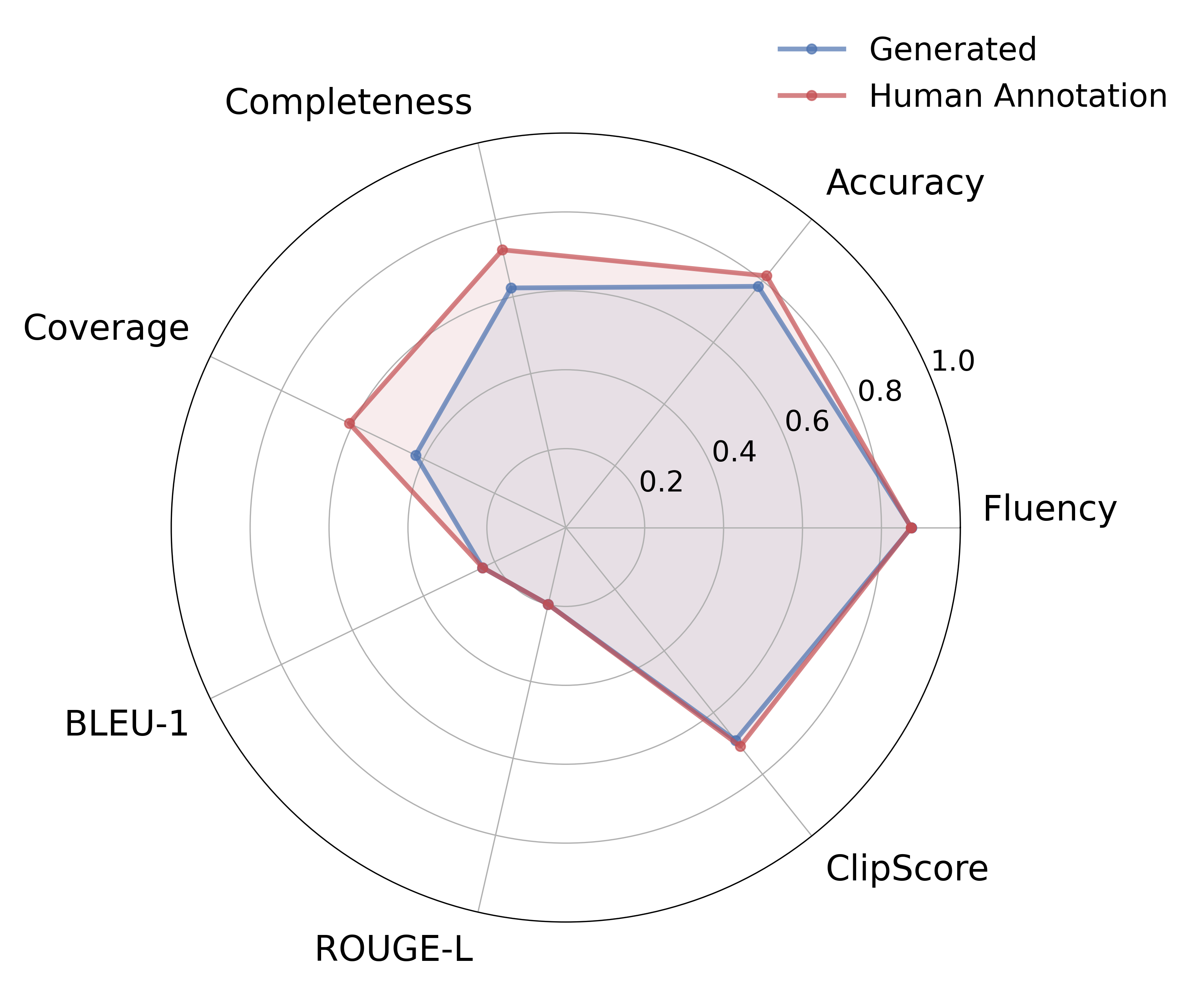}}
  \subcaptionbox{\label{fig:ablation}}
    {\includegraphics[width=0.25\linewidth]{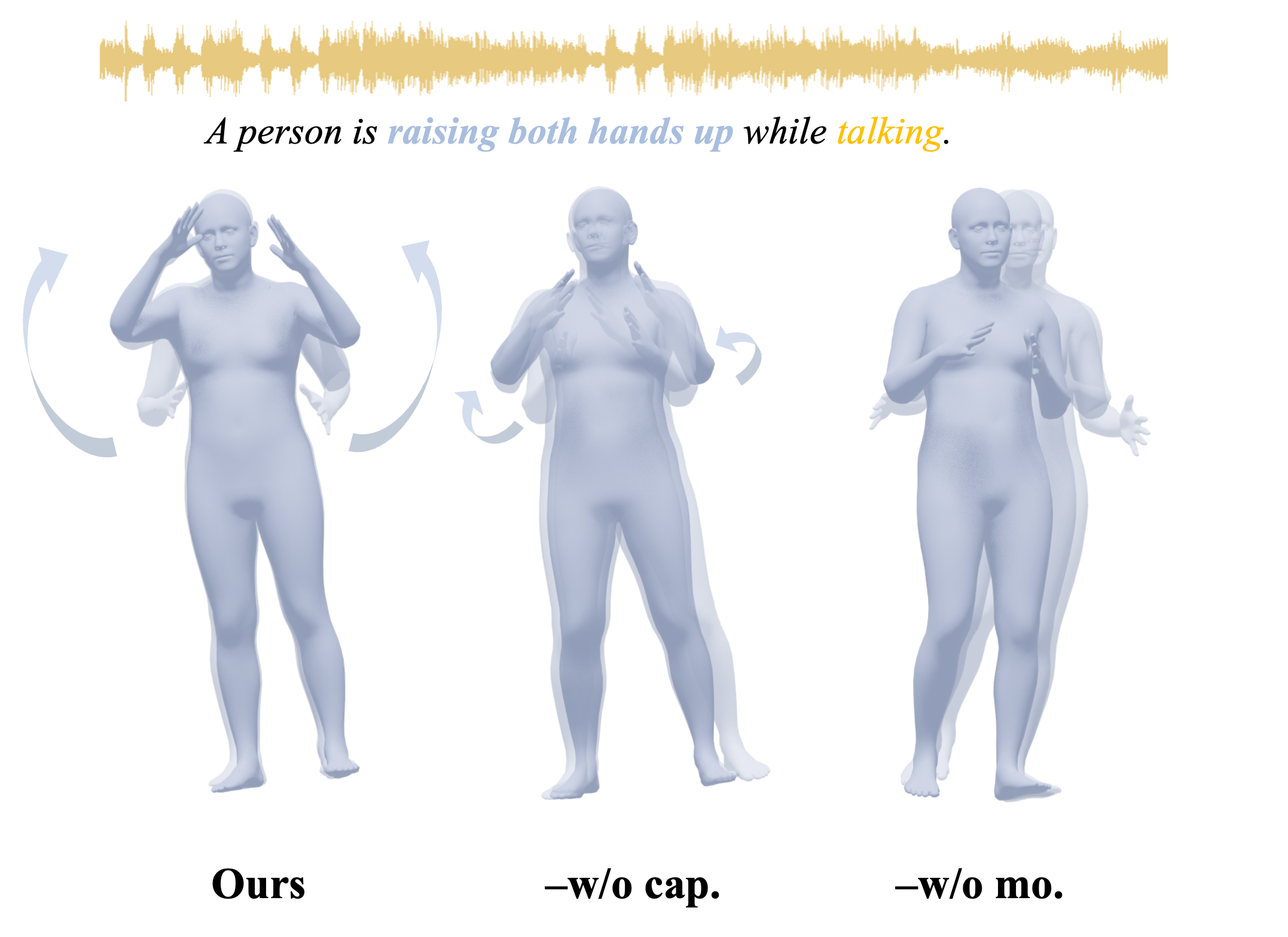}}
\vspace{-5pt}
    \caption{Visualization results. (a) Gesture captioning examples. Our captioning framework can effectively describe both overall motion patterns and fine-grained details. More results are in supplementary material (Sec.~\ref{sec:appendix_visual}). (b) Quantitative captioning evaluation. Our model performs comparably to human annotations. (c) Qualitative ablation study. Results are generated using audio and single caption.}
\end{figure*}

\subsection{Gesture Captioning}
\paragraph{Qualitative Results}
As shown in Fig.~\ref{fig:visualization_caption}, the proposed captioning framework demonstrates practical capabilities in generating descriptive gesture annotations. 
Our method can accurately describe both fine-grained hand movements (e.g., \textit{``moving both hands...''}) and coarse-grained full-body actions (e.g., \textit{``moves their waist...''}).
Furthermore, the proposed approach proves advantageous in capturing continuous composite actions within extended time windows (e.g., \textit{``a person [motion 1], then [motion 2] as [motion 3]''}). 
This demonstrates the effectiveness of our method in bridging the semantic prior gap in gesture datasets.
Nevertheless, we also observe that the model encounters challenges in perceiving temporal order in longer gesture sequences. More results and discussion of limitations are provided in supplementary material (Sec.~\ref{sec:appendix_visual}, \ref{sec:appendix_limitation}). 

\paragraph{Quantitative Captioning Evaluation}
Quantitative evaluation is inherently challenging due to the lack of Ground-Truth captions, precisely the ``semantic prior gap'' we aim to address.
To compensate, we construct a 200-sample \textit{expert-annotated set} and evaluate from three aspects:
(1) motion-semantic relevance (MotionClipScore~\cite{tevet2022motionclip}), (2) language quality (via GPT~\cite{openai2024gpt4omini}), (3) reference alignment (BLEU, ROUGE).
As shown in Fig.~\ref{fig:radar}, our model performs comparably to expert annotations in both relevance and quality, with slightly better fluency (4.380 vs. 4.375), comparable ClipScore (0.690 vs. 0.709), and reasonable alignment for the open-ended task. 
Combined with qualitative results (Fig.~\ref{fig:visualization_caption}), these further confirm the effectiveness and robustness of our captioning framework.

\subsection{Ablation Studies}
\label{sec:exp_ablation}
Ablation studies are conducted to evaluate the impact of different components. %under this task.
Qualitative results are shown in Fig.~\ref{fig:ablation}, while quantitative results are presented in Table.~\ref{tab:main_results}.

\paragraph{Multi-Granular Gesture Caption}
Fig.~\ref{fig:ablation} (middle) shows that removing gesture captions results in only basic co-speech gestures, lacking meaningful non-spontaneous movements.
Table~\ref{tab:main_results} further demonstrates that removing multi-granular captioning (\textit{-w/o mgc.}) significantly reduces semantic relevance, as evidenced by an increased MM-Dist and a decreased R-Precision. These highlight the crucial role of the multi-granular captioning in providing precise semantic guidance. 
A detailed comparison of the three captioning strategies (\textit{Reg.}, \textit{Dyn.}, \textit{Hie.}) is provided in supplementary material (Sec.~\ref{sec:appendix_exp}); the Hierarchical strategy with Regular local captions is adopted, as it achieves the best overall balance across all metrics. 

% \vspace{-1pt}
\paragraph{Hierarchically Controlled Denoiser}
Table~\ref{tab:main_results} shows that removing the hierarchically controlled denoiser (\textit{-w/o hcd.}) degrades reconstruction quality and weakens condition coordination, resulting in a stronger bias toward co-speech gestures (BC $0.583\uparrow$) while losing semantic relevance (MM-Dist $0.058\uparrow$). This underscores the importance of the hierarchical denoiser for multimodal coordination.

\vspace{-1pt}
\paragraph{Unified Cross-Dataset Motion Representation} 
Ablating unified motion representation (Fig.~\ref{fig:ablation} (right)) limits the semantic awareness beyond typical co-speech movements, leading to incomplete execution of intended movements like raising hands.
Results in Table~\ref{tab:main_results} (\textit{-w/o mo.}) further confirm this degradation, showing a substantial drop in semantic relevance (MM-Dist $0.850\uparrow$ and R-Precision $0.057\downarrow$).

\paragraph{Inference Time}
Long inference time is a major bottleneck in diffusion models.
We evaluate efficiency using Average Inference Time per Sentence (AITS)~\cite{chen2023executing} on a single NVIDIA RTX 3090 GPU, with batch size one and excluding model loading time. 
% As shown in Table.~\ref{tab:inference_time}, 
Our method achieves an AITS of $0.842^{\pm .002}$s, \textbf{over $6\times$ faster} than FreeTalker ($6.632^{\pm .044}$s) and SynTalker ($5.804^{\pm .044}$s).
This significant speedup can be attributed to our efficient hierarchical denoiser and optimized latent diffusion process, making our approach more practical for real-world applications.

%%%%%%%%%%%%%%%%%%%%%%%%%%%%%%%%%%%%%%%%%%%%%%%%%%%%%%%%%%%%%%%%%%%%%%%%%%%%%%%%%%%%%%%%%%%%%%%%%%

\vspace{6pt}
\section{Conclusion}
\vspace{3pt}
\label{sec:conclusion}
In this study, we propose CoordSpeaker, a coordinated gesture generation approach designed to generate both co-speech spontaneous gesture and caption-driven non-spontaneous motion, under joint audio-caption control.
We present the first Gesture Captioning Framework to bridge semantic prior gap by generating descriptive, well-aligned captions for gesture data. 
Building upon this, we propose a Coordinated Gesture Generation Model, leveraging a hierarchically controlled denoiser for efficient and coordinated gesture generation. 
Our approach pioneers gesture captioning and explores bidirectional gesture-text mapping. 
Extensive experiments demonstrate that CoordSpeaker produces high-quality gestures with enhanced semantic coherence and rhythmic synchronization while significantly improving efficiency, surpassing existing methods.

\clearpage

{
    \small
    \bibliographystyle{ieeenat_fullname}
    \bibliography{main}
}

\clearpage
\setcounter{page}{1}
\maketitlesupplementary

In the supplementary material, we provide more implementation details (Sec.~\ref{sec:appendix_imple}), additional experimental results (Sec.~\ref{sec:appendix_exp}), more visual results (Sec.~\ref{sec:appendix_visual}), discussion on limitations and future work (Sec.~\ref{sec:appendix_limitation}), and user study ethical considerations (Sec.~\ref{sec:appendix_userstudy}) of the proposed CoordSpeaker.

\section{Implementation Details}
\label{sec:appendix_imple}
\subsection{Network Details}
Our transformer-based VAE and denoiser $\epsilon_\theta$ are both composed of an encoder and a decoder, each containing 9 layers and 4 attention heads with GELU activation and residual connections. The latent dimension is set to $z \in \mathbb{R}^{1 \times 512}$.  
For training, we use the AdamW optimizer with a learning rate of $1e^{-4}$ and a batch size of 128. The VAE is trained for 6000 epochs, while the diffusion model is trained for 2000 epochs. 
We use 1000 diffusion steps for training and 50 steps for inference.
During training, the noise variance $\beta_t$ linearly scaled from $8.5 \times 10^{-4}$ to 0.012. In the VAE stage, the KL loss weight $\beta$ is set to $1e^{-4}$. %~\ref{eq:loss_vae}).  
During inference, for classifier-free guidance, the audio and caption guidance scales are set to $s_1=7$ and $s_2=0.75$ by default to balance contributions.

For condition embedding, we employ CLIP text encoder~\cite{radford2021clip} and WavLM encoder~\cite{chen2022wavlm} to extract semantic features $\mathbf{C} \in \mathbb{R}^{512}$ from captions and audio features $\mathbf{A} \in \mathbb{R}^{T\times1133}$ from speech respectively. 
Both semantic and audio embeddings are projected through a linear layer into a 512-dimensional space before being fed into the denoiser. 

\subsection{Additional Details of Gesture Captioning}
\label{sec:appendix_cap}

\paragraph{Prompt Template}
Table~\ref{tab:prompt_templates} presents a collection of prompt templates employed in our gesture captioning framework. These carefully curated templates are randomly sampled multiple times and paired with different gesture segments to generate diverse gesture captions. Templates are inspired by recent advances in motion-language modeling~\cite{jiang2024motiongpt}.

\paragraph{Motion Representation Alignment} % Conversion
To better match the general motion-language space, we convert the gesture features into a commonly adopted human motion format~\cite{guo2022humanml3d} $x\in \mathbb{R}^{T\times 659} \rightarrow \mathbb{R}^{T\times 263}$ by retaining 22 key joints before performing captioning inference. Nevertheless, this conversion may omit some fine-grained finger-motion details. We provide more discussions in the following section (Sec.~\ref{sec:appendix_limitation}). 

\paragraph{Caption Quality Control}
Given the differences in granularity and distribution between full-body motion and co-speech gestures, MotionLLM, pretrained on coarser human motion data, occasionally generate overly brief or action-oriented descriptions, such as interpreting exaggerated speaker movements as ``The person is boxing''. 
To mitigate this, we implement a quality control mechanism that filters out captions with fewer than 5 words and their corresponding gesture segments, which are considered to lack clear non-spontaneous motion and cannot provide sufficient semantic guidance. 
This mechanism ensures that each retained segment is paired with a semantically rich caption as an effective training prior. 
In addition, global captions further complement local ones by providing broader contextual semantics. This strategy ensures caption quality and reduces the risk of ambiguous guidance during generation.

\begin{table*}[t]
\centering
\caption{Examples of prompt templates used in our gesture captioning framework.}
\label{tab:prompt_templates}
\begin{tabular}{llc}
\toprule
Task & Input & Output \\
\midrule
\multirow{10}{*}{Gesture-to-Text} & Give me a summary of the motion being displayed in [motion] using words. & \multirow{10}{*}{[caption]} \\
& Explain the motion illustrated in [motion] using language. & \\
& Describe the action being represented by [motion] using text. & \\
& What kind of action is being demonstrated in [motion]? & \\
& Describe the movement demonstrated in [motion] in words. & \\
& Generate a sentence that explains the action in [motion]. & \\
& Please describe the movement depicted in [motion] using natural language. & \\
& Provide a description of the motion being displayed in [motion] using language. & \\
& Give me a brief summary of the movement depicted in [motion]. & \\
& Describe the movement demonstrated in [motion] using natural language. & \\
\bottomrule
\end{tabular}
\end{table*}

\begin{table*}[t]
    \centering
    \caption{Ablation studies on multi-granular captioning strategies. ``Reg." denotes the Regular Caption strategy, ``Dyn." denotes the Dynamic Caption strategy, and ``Hie." denotes the Hierarchical Caption strategy.
    Each metric is reported under the 95\% confidence interval from 20 times running. We report BC $\times 10^{-1}$ and Top-1 R-Precision.}
    \small
    \label{tab:ablation}
    \begin{tabular}{l c@{\hspace{3pt}} c@{\hspace{3pt}} c@{\hspace{3pt}} c@{\hspace{3pt}} c@{\hspace{3pt}} c@{\hspace{3pt}} c@{\hspace{3pt}} c@{\hspace{3pt}} c}
    \toprule
    \multirow{3}{*}{Methods} & \multicolumn{2}{c}{Reconstruction} & \multicolumn{3}{c}{Audio-to-Gesture} & \multicolumn{4}{c}{Text-to-Motion} \\
    \cmidrule(lr){2-3} \cmidrule(lr){4-6} \cmidrule(lr){7-10}
    & Jerk$\rightarrow$ & Accel.$\rightarrow$ & FGD$\downarrow$ & BC$\uparrow$ & L1Div$\uparrow$ & FID$\downarrow$ & MM-Dist$\downarrow$ & Div$\rightarrow$ & R-Precision$\uparrow$ \\
    \midrule
    GT & $1.165^{\pm .000}$ & $0.043^{\pm .000}$ & - & - & - & - & $6.205^{\pm .043}$ & $5.512^{\pm .114}$ & $0.140^{\pm .008}$  \\
    \midrule
    Ours-Reg.  & $1.201^{\pm .017}$ & $0.038^{\pm .001}$ & $2.302^{\pm .061}$ & $1.910^{\pm .004}$ & $12.781^{\pm .044}$ & $1.260^{\pm .063}$ & $6.872^{\pm .058}$ & $5.303^{\pm .107}$ & $0.102^{\pm .010}$  \\
    Ours-Dyn.  & $1.189^{\pm .013}$ & $0.038^{\pm .000}$ & $2.866^{\pm .106}$ & $1.943^{\pm .037}$ & $14.471^{\pm .110}$ & $1.404^{\pm .049}$ & $6.955^{\pm .044}$ & $5.440^{\pm .114}$ & $0.095^{\pm .006}$  \\
    Ours-Hie.  & $1.190^{\pm .015}$ & $0.039^{\pm .001}$ & $3.173^{\pm .123}$ & $1.327^{\pm .049}$ & $10.861^{\pm .066}$ & $1.118^{\pm .061}$ & $6.814^{\pm .056}$ & $5.558^{\pm .126}$ & $0.100^{\pm .008}$ \\
    \bottomrule
    \end{tabular}
\end{table*}

\subsection{Dataset Details}
To balance the data distribution between datasets during training, a weighted random sampling strategy is employed for the dataloader.
Following~\cite{yang2024freetalker}, all motion sequences are resampled to 20 FPS and either truncated or padded to 180 frames. For the HumanML3D dataset, only sequences with lengths between 40 and 180 frames are utilized. Data is split into training, validation, and testing sets in an 8:1:1 ratio.

\subsection{Evaluation Metrics}
\label{sec:appendix_metric}

\paragraph{Coordination Evaluation Protocol}
Due to the absence of a unified multimodal benchmark, we follow standard practice~\cite{yang2024freetalker,chen2024syntalker} and report Audio-to-Gesture and Text-to-Motion metrics on BEAT and HumanML3D in Sec.~\ref{sec:exp_main}, respectively, to ensure fair comparison. 
However, this separation forces existing quantitative metrics to evaluate only single-modality factors: speech–gesture synchrony on BEAT and text–motion semantic alignment on HumanML3D, without directly reflecting the multimodal coordination that is critical to the joint generation task.
Consequently, in Table~\ref{tab:main_results} we focus on balanced performance across Audio-to-Gesture and Text-to-Motion metrics, as strong multimodal coordination inherently requires trade-offs between separate tasks. 
We believe that developing a unified multimodal benchmark would substantially benefit the coordination evaluation of this field, and our captioning framework may help facilitate its construction.

\paragraph{Fréchet Gesture Distance}
Following prior work~\cite{liu2024emage}, our FGD is calculated based on latent features extracted by a pre-trained autoencoder. Specifically, FGD is computed as:
\begin{equation}
    \text{FGD}(g, \hat{g}) = \|\mu_r - \mu_g\|_2^2 + \text{Tr}(\Sigma_r + \Sigma_g - 2(\Sigma_r\Sigma_g)^{1/2})
\end{equation}
where $\mu_r$ and $\Sigma_r$ represent the mean and covariance matrix of the latent features $z_r$ extracted from real gestures $g$, while $\mu_g$ and $\Sigma_g$ correspond to those of generated gestures $\hat{g}$. A lower FGD indicates a better quality of generated gestures.
To extract these latent features for our proposed unified motion representation (Sec. 3.1), %(Sec.~\ref{sec:method:m_rep}), 
we train a Full CNN-based autoencoder consisting of 4-layer convolutional encoders and decoders. Each convolutional layer is followed by a LeakyReLU activation function. The latent dimension is set to 240. This autoencoder is trained on the 4 English speakers of the BEAT dataset for 1000 epochs using the same training configuration as our VAE stage.

\section{Additional Experimental Results}
\label{sec:appendix_exp}
\subsection{Comparison of Multi-Granular Captioning}
\label{sec:appendix_exp_mgc}
Table~\ref{tab:ablation} further compares the performance of different captioning strategies.
Among these, the hierarchical strategy (\textit{Ours-Hie.}) achieves the optimal balance across all metrics, making it well-suited for coordinated gesture generation. 
It yields better semantic relevance (lower MM-Dist at $6.814$ vs. $6.872$ and $6.955$), improved motion quality (lower FID by 11.3\% and 25.6\%), while maintaining comparable audio synchronization and motion diversity.
Additionally, the dynamic strategy (\textit{Ours-Dyn.}) exhibits advantages in co-speech gesture synchronization (BC: $1.943$) and diversity (L1Div: $14.471$). This may be attributed to its adaptive sampling mechanism, which introduces more rhythmic variation during training. 
Overall, these results suggest that the multi-granular captioning mechanism effectively supports multimodal coordination, enabling both fine-grained semantic control and rhythmically natural gestures.

\section{More Visual Results} 
\label{sec:appendix_visual}
\subsection{More Coordinated Generation Results}
\begin{figure*}[t]
    \centering
    \includegraphics[width=0.85\linewidth]{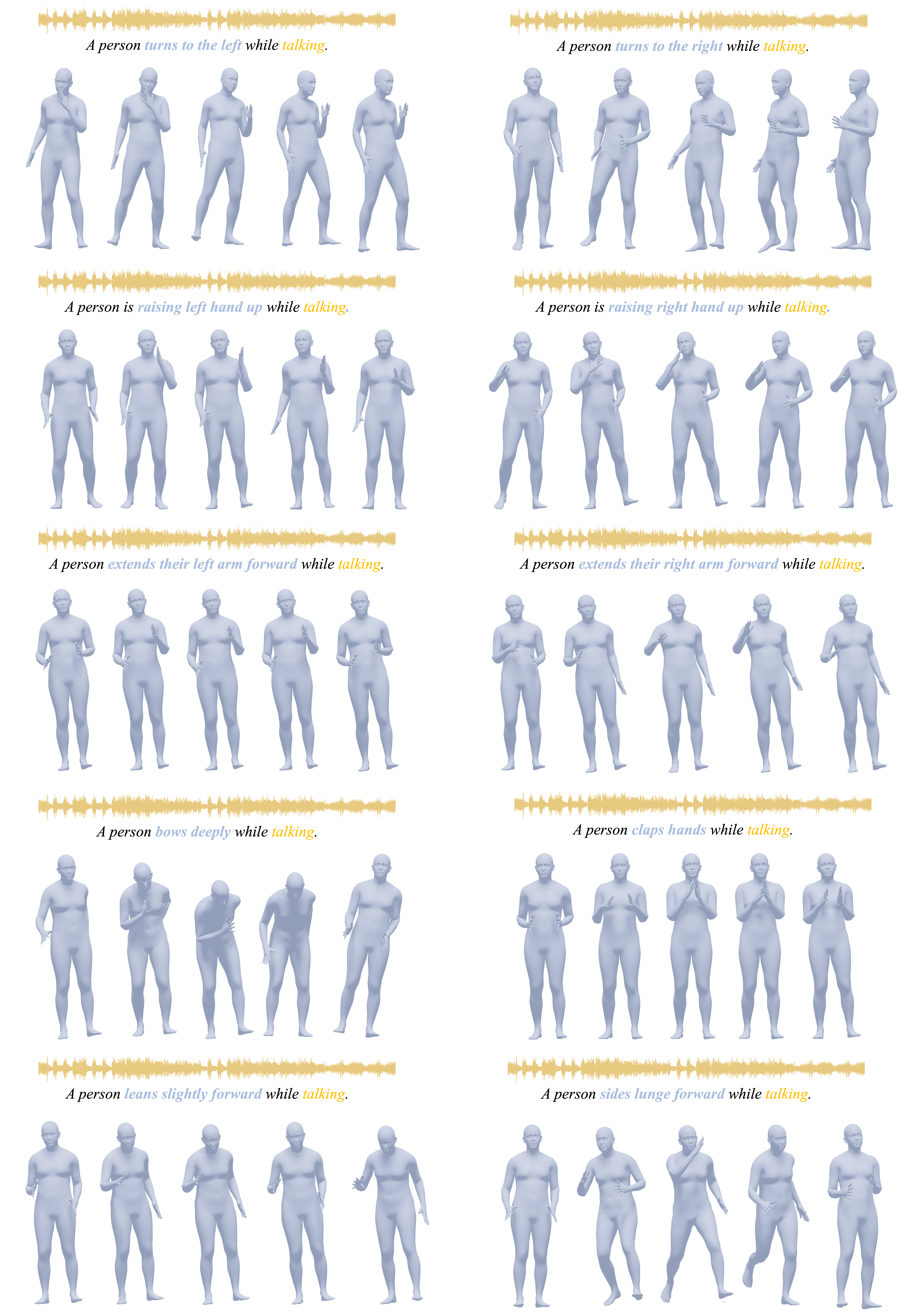}
    \caption{More visual results of coordinated gesture generation.}
    \label{fig:vis_supply}
\end{figure*}

As shown in Fig.~\ref{fig:vis_supply}, we provide more coordinated generation results using \textit{calm} audio and different text captions.
These results further confirm the effectiveness of our proposed method in generating both co-speech spontaneous gestures and caption-driven non-spontaneous motions under joint speech-caption control. For dynamic results, please see our demo video appendix.

\subsection{More Gesture Captioning Results}
\begin{figure*}[t]
    \centering
    \includegraphics[width=0.85\linewidth]{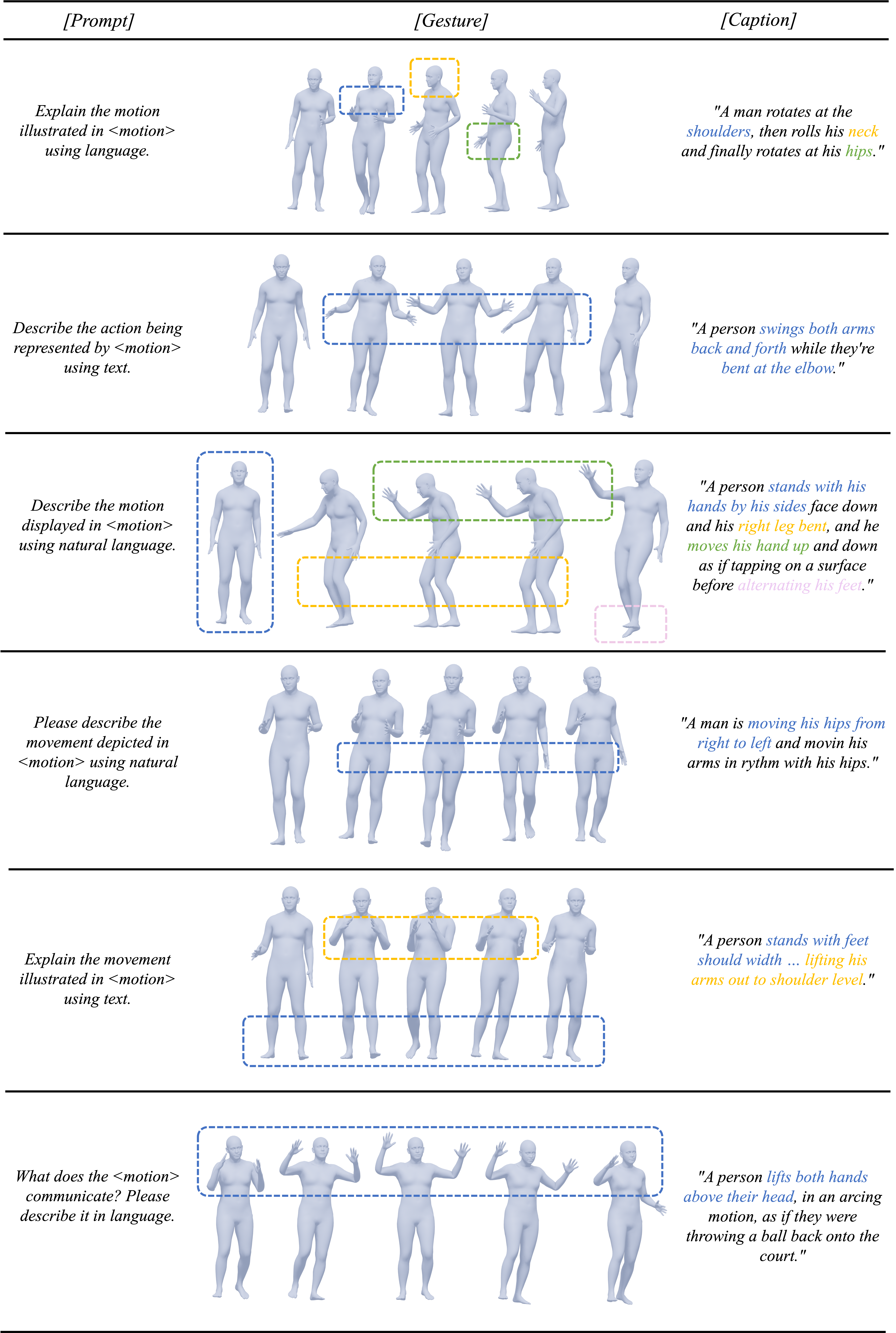}
    \caption{More gesture captioning results. \textcolor[RGB]{100,149,237}{Co}\textcolor[RGB]{204,153,0}{lo}\textcolor[RGB]{34,139,34}{rf}\textcolor[RGB]{255,182,193}{ul} boxes highlight the precise mapping between gestures and textual captions.}
    \label{fig:vis_caption_supply}
\end{figure*}

We present additional gesture captioning results in Fig.~\ref{fig:vis_caption_supply}, further demonstrating the effectiveness of our approach in accurately mapping gestures to text. 
As highlighted in colorful boxes, the model effectively captures both fine-grained hand movements and coarse-grained full-body motions while describing complex, continuous actions.

\section{Limitations and Future Work} 
\label{sec:appendix_limitation}
\subsection{Enhancing Gesture Understanding}
While gesture captioning effectively bridges the semantic gap in gesture generation, it still faces challenges in temporal consistency, occasionally leading to misordered actions in longer sequences. 
Our multi-granular captioning mechanism mitigates this: fine-grained local captions reduce the burden of describing long sequences, while global captions provide complementary long-range semantic context. Future improvements may stem from enhancing the temporal modeling capacity of motion-language models.

In addition, since MotionLLM occasionally produces coarse action-oriented descriptions, constructing broader gesture–caption benchmarks and fine-tuning a dedicated GestureLLM 
could further enhance the perception of fine-grained gesture semantics.
We aim to expand our expert-annotated set in future work to support this direction.
Moreover, incorporating audio or text transcripts into caption generation offers a promising avenue for producing more expressive gesture captions, which could further enhance coordinated gesture generation.

\subsection{Fine-grained Gesture Representations}
The proposed coordinated gesture generation framework could benefit from more refined motion representations. Given our primary focus on coordinated gestures and full-body movement synthesis, we adopt the BEAT dataset~\cite{liu2022beat}, which provides sufficient data for this purpose. However, integrating datasets with more precise head and finger motion, such as BEAT2~\cite{liu2024emage}, could facilitate more holistic gesture generation. A key challenge that lies here is bridging the additional semantic gap for finer-grained facial expressions and finger movements, potentially requiring more detailed annotated datasets or a more powerful gesture-language model in future research.

\section{Ethical Considerations in User Study}
\label{sec:appendix_userstudy}
\begin{figure*}[t]
    \centering
    \includegraphics[width=\linewidth]{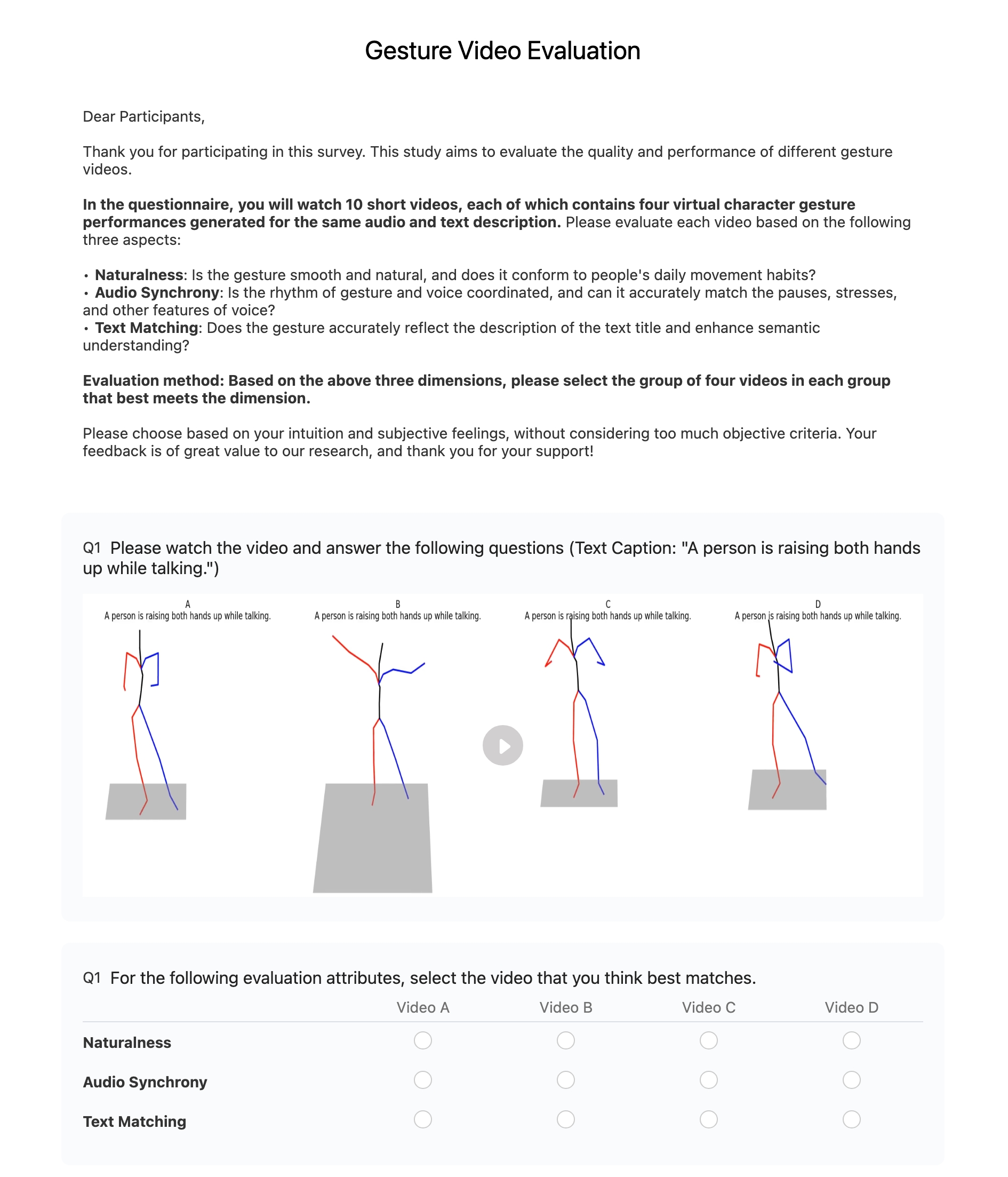}
    \caption{The screenshots of the user study website for participants.}
    \label{fig:user_study}
\end{figure*}

This section elaborates on the details of our user study protocol and participant demographics. The study recruited participants between 18 and 40 years of age, all possessing a minimum of an undergraduate degree to ensure a qualified assessment. Fig.~\ref{fig:user_study} illustrates the interface of our evaluation platform, which presents a standardized template layout to all participants. To maintain data quality and ensure thorough evaluation, we implemented a response time threshold: any trial completed in less than 100 seconds was deemed insufficient for proper assessment and subsequently excluded from our analysis.

\end{document}